# Ontology as a Source for Rule Generation

Olegs Verhodubs
Riga Technical University
Riga, Latvia
Email: oleg.verhodub@inbox.lv

## ABSTRACT

This paper discloses the potential of OWL (Web Ontology Language) ontologies for generation of rules. The main purpose of this paper is to identify new types of rules, which may be generated from OWL ontologies. Rules, generated from OWL ontologies, are necessary for the functioning of the Semantic Web Expert System. It is expected that the Semantic Web Expert System (SWES) will be able to process ontologies from the Web with the purpose to supplement or even to develop its knowledge base.

## I INTRODUCTION

The Web changes the way people communicate with each other, and it lies at the heart of a revolution that is currently transforming the developed world toward a knowledge society [1]. Today the Web is used for seeking and making use of information, searching for and getting in touch with other people, reviewing catalogs of online stores and ordering products by filling out forms. These activities are performed by the user, and they are not supported by software tools particularly well. Keyword-based search engines, which are the main software tools for these activities, have several serious problems. The first problem is low precision. If the main relevant pages are retrieved, they are of little use if another 30,000 mildly relevant or irrelevant documents are also retrieved. The next problem is that results are sensitive to vocabulary. Initial keywords do not get the results we want. The third problem is that results are single Web pages. Information is spread over a lot of documents, and it is necessary to initiate several queries to collect relevant documents. After that partial information has to be manually extracted and put together. And despite the growing quality of keyword-based search engines, someone needs to browse selected documents and extract the information [1].

Another approach is based on the use of the Semantic Web technologies. It is more machine – processable, and the fundamental principle of this approach is to utilize semantic metadata [2]. Semantic metadata may describe a document or part of a document. They also may describe entities within the document. Thus, the metadata is semantic, that is, it tells about the content of a document. This differs from the today's Web, encoded in HTML (HyperText Markup Language), which purely describes the format in which the information should be presented [2], and thus the content of the today's Web is formatted for human readers rather than programs.

At the heart of all Semantic Web applications is the use of ontologies. Ontologies are an expression of semantic metadata. Ontology formally describes a domain of interest. It consists of a finite list of terms and the relationships between them. OWL specification endorsed by W3C (World Wide Web Consortium), and it is intended for ontology development [2].

Ontologies may be useful not only to specify terms and relationships between them, that is, to represent knowledge. It was concluded that it was possible to generate rules from OWL ontologies [3]. There were investigated several cases when OWL ontology code fragments could be transformed to rules. But investigated cases were not exhaustive. In this connection the main purpose of this paper is to identify new types of rules, which may be generated from the OWL ontology, in order to turn it into full-fledged and self-sufficient resource for rule generation. Generated rules are necessary for construction or supplementation of the Semantic Web Expert System (SWES) knowledge base.

The final goal of the research is to develop the SWES. SWES is a new expert system, which will be capable to use OWL ontologies from the Web, to generate rules from them and to supplement its knowledge base in automatic mode [4].

The paper is organized as follows. The next section shows OWL code fragments, which can be transformed to rules. Section III gives a classification of generated rules. The last section presents conclusions and ideas for future work.

## II RULE GENERATION FROM ONTOLOGIES

There are several languages for coding rules. They are RuleML (Rule Markup Language), R2ML (REWERSE Rule Markup Language), SWRL (Semantic Web Rule Language), RIF (Rule Interchange Format). Rules in these languages are defined by the user directly. But when we talk about the task of rule generation from OWL ontologies, we mean "net" OWL ontology. Here "net" OWL ontology means ontology without using any rule languages. The basic idea is to determine OWL ontology code fragments, which can be transformed to rules [3]. This idea had already been presented, and there were described several cases when OWL ontology code fragments could be transformed to rules [3]. However it is possible to investigate some other cases, when OWL ontology code fragments can be transformed to rules. Let us take them in order.

When a class has properties, it is possible to generate a rule. For instance, there is "Car" class with two properties "Wheel" and "Engine" (Fig. 1):

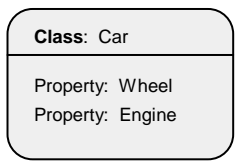

Fig. 1. "Car" class with two properties.

It is possible to generate the following rule:

**IF** Car **THEN** Wheel **and** Engine (1)

This rule can be generated from the following OWL code fragment:

```
<owl:Classrdf:ID="#Car "/>
```

```
<owl:DatatypePropertyrdf:ID="Wheel">
<rdfs:domainrdf:resource="#Car "/>
<rdfs:rangerdf:resource="xs:string"/>
</owl:DatatypeProperty>
```

```
<owl:DatatypePropertyrdf:ID="Engine">
<rdfs:domainrdf:resource="#Car "/>
<rdfs:rangerdf:resource="xs:string"/>
</owl:DatatypeProperty>
```

Another case is when ontology has two equivalent classes. For example, there are "Car" and "Auto"classes. Class "Auto"is equivalent to class "Car". Class "Car" has "part of" relation to class "Vehicle" (Fig. 2.):

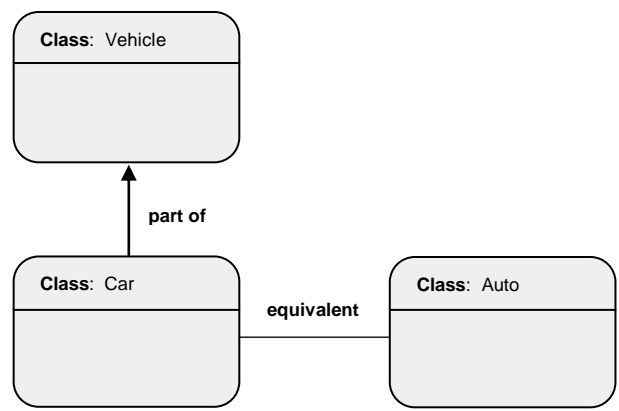

Fig. 2.Two equivalent classes "Car" and "Auto".

It is possible to generate the following rule:

**IF**Car**equivalent** Auto **THEN** ("part of" Vehicle) ∈Auto (2)

This rule can be generated from the following OWL code fragment:

```
<owl:Classrdf:ID="Auto">
<owl:equivalentClass>
<owl:Classrdf:ID="Car"/>
</owl:equivalentClass>
</owl:Class>
<owl:Classrdf:ID="Car">
<rdfs:subClassOfrdf:resource="#Vehicle"/>
</owl:Class>
-------------------------- OR ----------------------
<owl:Classrdf:ID="Auto">
<owl:sameAsrdf:resource="# Car"/>
</owl:Class>
<owl:Classrdf:ID="Car">
<rdfs:subClassOfrdf:resource="#Vehicle"/>
</owl:Class>
```

In the case, when there is a relation between two classes, it is also possible to generate a rule. For example, there are "Man" and "House" classes and also "liveIn" relation between these two classes (Fig.3.):

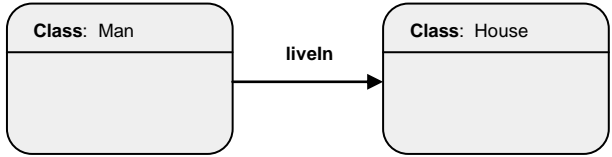

Fig. 3. Two classes "Man" and "House" has relation "liveIn" between them.

It is possible to generate such rule:

**IF** (liveIn House) **THEN** Man, (3)

Let us explain the rule. This rule means that if there is some instance which "liveIn House" then this instance belongs to class "Man". This rule can be generated from the following OWL code fragment:

```
<owl:ObjectPropertyrdf:ID="liveIn">
<rdfs:domainrdf:resource="#Man"/>
<rdfs:rangerdf:resource="#House"/>
</owl:ObjectProperty>
```

The next case is the case when there are three classes "House", "City", "Country" and there is "part of" relation between the "House" and the "City" classes, and also there is "part of" relation between "City" and "Country" classes (Fig. 4):

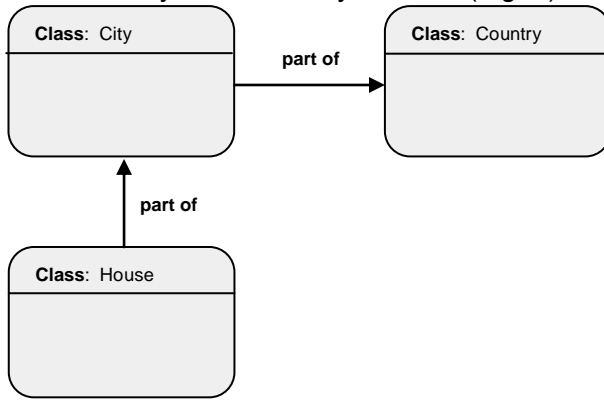

Fig. 4. "House" class is part of "City" class and "City" class is part of "Country" class.

This rule can be generated as follows:

**IF** (House "part of" City) **and** (City "part of" Country)
**THEN** (House "part of" Country)(4)

The rule means that if there is the first class, which is part of the second class, and the second class is part of the third class then the first class is part of the third class. Such rule can be generated from the following OWL ontology code fragment:

```
<owl:Classrdf:ID="House">
<rdfs:subClassOfrdf:resource="#City"/>
</owl:Class>
<owl:Classrdf:ID="City">
<rdfs:subClassOfrdf:resource="#Country"/>
</owl:Class>
-------------------- OR ---------------------
<owl:Classrdf:ID="House">
<rdfs:subClassOf>
<owl:Classrdf:ID="City"/>
</rdfs:subClassOf>
</owl:Class>
<owl:Classrdf:ID="City">
<rdfs:subClassOf>
<owl:Classrdf:ID="Country"/>
</rdfs:subClassOf>
</owl:Class>
---------------------- OR ---------------------
<owl:Classrdf:ID="House">
<rdfs:subClassOf>
<owl:Classrdf:about="#City"/>
</rdfs:subClassOf>
</owl:Class>
<owl:Classrdf:ID="City">
<rdfs:subClassOf>
<owl:Classrdf:about="#Country"/>
</rdfs:subClassOf>
</owl:Class>
```

One more case which can provide with rules is when there are three classes and two relations between them. One of these relations is "part of" relation. For instance, there are "Man", "House", "City" classes and "liveIn" relation between "Man" and "House" classes. There also is "part of" relation between "House" and "City" classes (Fig.5):

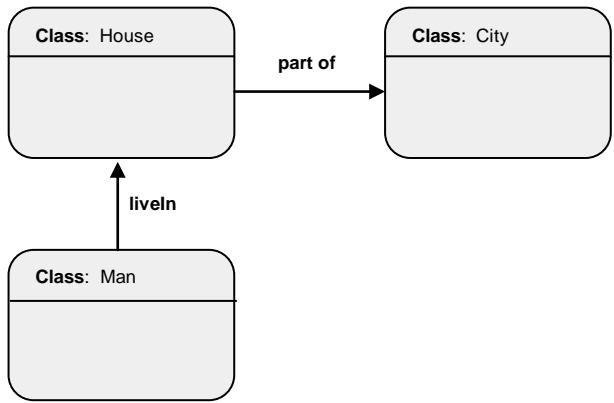

Fig. 5. "Man", "House", "City" classes and "liveIn", "part of" relations between them.

It is possible to generate the rule:

**IF** (Man "liveIn" House) **and** (House "part of" City)

**THEN** (Man "liveIn" City),(5)

This rule means that if there are three classes, where there is some relation between the first and the second classes and there is "part of" relation between the second and the third classes, then there is relation between the first and the third classes such as between the first and the second classes. This rule can be generated from the following OWL ontology code fragment:

```
<owl:Classrdf:ID="House">
<rdfs:subClassOfrdf:resource="#City"/>
</owl:Class>
<owl:ObjectPropertyrdf:ID="liveIn">
<rdfs:domainrdf:resource="#Man"/>
<rdfs:rangerdf:resource="#House"/>
</owl:ObjectProperty>
-------------------- OR --------------------
<owl:Classrdf:ID="House">
<rdfs:subClassOf>
<owl:Classrdf:ID="City"/>
</rdfs:subClassOf>
</owl:Class>
<owl:ObjectPropertyrdf:ID="liveIn">
<rdfs:domainrdf:resource="#Man"/>
<rdfs:rangerdf:resource="#House"/>
</owl:ObjectProperty>
--------------------- OR ---------------------
<owl:Classrdf:ID="House">
<rdfs:subClassOf>
<owl:Classrdf:about="#City"/>
</rdfs:subClassOf>
</owl:Class>
<owl:ObjectPropertyrdf:ID="liveIn">
<rdfs:domainrdf:resource="#Man"/>
```

<rdfs:rangerdf:resource="#House"/>
</owl:ObjectProperty>

The following case is when two properties are defined and one of them is the subproperty of another one. For example, there is a property "hasFather", which is the subproperty of property "hasParent" (Fig. 6.):

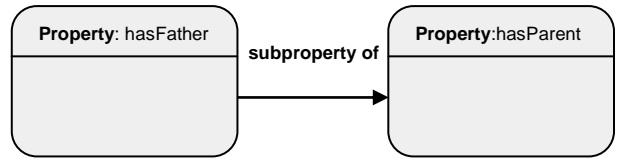

Fig. 6. Property "hasFather" is subproperty of property "hasParent".

It is possible to generate the following rule:

**IF** hasFather **and** "subproperty of" **THEN** hasParent, (6)

Let us explain this rule. The rule means that if there is an instance of property "hasFather" and the property "hasFather" is the subproperty of property "hasParent", then this instance belongs to the property "hasParent". This rule can be generated from the following OWL code fragments:

```
<owl:ObjectPropertyrdf:ID="hasFather">
<rdfs:subPropertyOfrdf:resource="#hasParent"/>
</owl:ObjectProperty>
```

When ontology has a symmetric property it is possible to generate several rules. For example, there are "Programmer" and "Engineer" classes and also the symmetric property "colleagueOf" between these classes (Fig.7.):

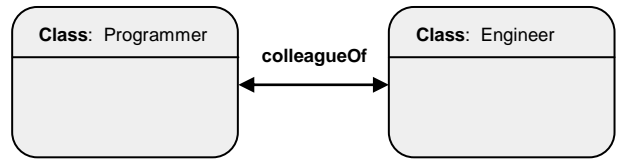

Fig. 7.Symmetric property "colleagueOf" between "Programmer" and "Engineer" classes.

It is possible to generate such rules:

**IF** Programmer **THEN** (colleagueOf Engineer), (7)

**IF** Engineer **THEN** (colleagueOf Programmer), (8)

The first rule means that if there is an instance of class "Programmer" then this instance has relation "colleagueOf" to the class "Engineer". The second rule means that if there is an instance of class "Engineer" then this instance has relation "colleagueOf" to the class "Programmer". These rules can be generated from the following OWL code fragments:

```
<owl:SymmetricPropertyrdf:ID="colleagueOf">
<rdfs:domainrdf:resource="#Programmer"/>
<rdfs:rangerdf:resource="#Engineer"/>
</owl:SymmetricProperty>
```

The next case of rule generation is when ontology has a transitive property. For instance, there is a transitive property "subAreaOf" between "Latgale", "Latvia" and "EU" classes (Fig. 8.):

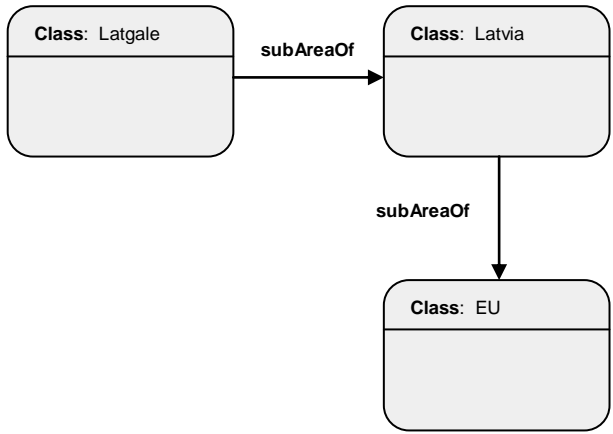

Fig. 8. Transitive property "subAreaOf" between three classes.

It is possible to generate the rule:

**IF** (Latgale "subAreaOf" Latvia) **and** (Latvia "subAreaOf" EU)

**THEN** (Latgale "subAreaOf" EU),  (9)

This rule means that if, for example, there are "Latgale", "Latvia", "EU" classes, and there is "subAreaOf" transitive relation between "Latgale" and "Latvia" and also between "Latvia" and "EU" classes, then there is "subAreaOf" relation between "Latgale" and "EU" classes. This rule can be generated from the following OWL code fragments:

```
<owl:Classrdf:ID="Latgale">
<subAreaOf>
<owl:Classrdf:ID="Latvia">
</subAreaOf>
</owl:Class>
<owl:Classrdf:ID="EU"/>
<owl:Classrdf:about="#Latvia">
<subAreaOfrdf:resource="#EU"/>
</owl:Class>
<owl:TransitivePropertyrdf:ID="subAreaOf">
```

```
<rdf:typerdf:resource=http://www.w3.org/2002/07/owl#ObjectProperty/>
</owl:TransitiveProperty>
---------------------- OR ----------------------
<owl:Classrdf:ID="Latgale">
<subAreaOfrdf:resource="#Latvia">
</owl:Class>
<owl:Classrdf:ID="EU"/>
<owl:Classrdf:ID="Latvia">
<subAreaOfrdf:resource="#EU"/>
</owl:Class>
<owl:TransitivePropertyrdf:ID="subAreaOf">
<rdf:typerdf:resource=http://www.w3.org/2002/07/owl#ObjectProperty/>
</owl:TransitiveProperty>
```

When ontology has a class, which has only one "partOf" relation, then it is possible to generate rule. For instance, there is "City" class, and it has only one "partOf" relation (Fig.9.):

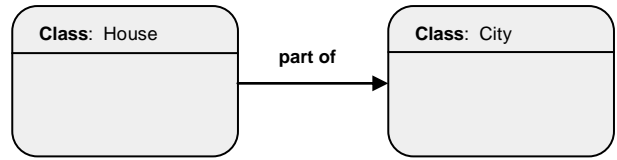

Fig. 9. The only one "part of" relation of class "City".

It is possible to generate the following rule:

**IF** City **and** only one "part of"  **THEN** (more "part of" ∈City),  (10)

Let us explain this rule. If there is a class "City", which has only one "part of" relation, then it has one or more "part of" relations, too. That is, the "City" class has not only one "House" part, but one or more other parts. This rule can be generated from the following OWL code fragments:

```
<owl:Classrdf:ID="House">
<rdfs:subClassOfrdf:resource="#City"/>
</owl:Class>
-------------------- OR --------------------
<owl:Classrdf:ID="House">
<rdfs:subClassOf>
<owl:Classrdf:ID="City"/>
</rdfs:subClassOf>
</owl:Class>
---------------------- OR ----------------------
<owl:Classrdf:ID="House">
<rdfs:subClassOf>
<owl:Classrdf:about="#City"/>
</rdfs:subClassOf>
```

```
</owl:Class>
```

In the case, when there is a relation between two classes, it is also possible to generate a rule. For example, there are "Fox" and "Hole" classes and also "liveIn" relation between these two classes (Fig.10.):

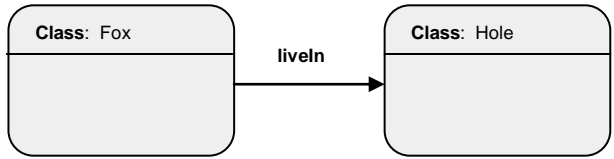

Fig. 10.Classes "Fox", "Hole" and relation "liveIn" between them.

It is possible to generate this rule:

**IF**Fox**and** Hole **THEN** liveIn, (11)

The rule means that if there are instances of "Fox" and "Hole" classes, then there is "liveIn" relation between these instances. This rule can be generated from the following OWL code fragment:

```
<owl:ObjectPropertyrdf:ID="liveIn">
<rdfs:domainrdf:resource="#Fox"/>
<rdfs:rangerdf:resource="#Hole"/>
</owl:ObjectProperty>
```

The following case is when ontology has a restriction with the value constraint "owl:allValuesFrom". For instance, there is a "hasPass" restriction with "owl:allValuesFrom" value constraint, which equals to the class "Citizen" (Fig. 11.):

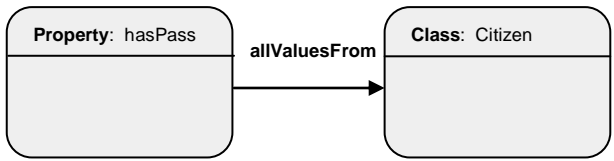

Fig. 11. "hasPass" property has values of class "Citizen", only.

It is possible to generate the following rule:

**IF**not Citizen **THEN** **not** hasPass, (12)

Let us explain the rule. If there is some instance, which does not belong to the "Citizen" class, then this instance does not apply to the property "hasPass". The rule can be generated from the following OWL ontology code fragment:

```
<owl:Restriction>
<owl:onPropertyrdf:resource="#hasPass" />
```

```
<owl:allValuesFromrdf:resource="#Citizen"  />
</owl:Restriction>
```

In the case when there are three classes and one of them is an intersection of other two classes, it is possible to generate rule. For example, there are "Human", "Man" and "Male" classes. The class "Man" is intersection of the "Human" and "Male" classes (Fig.12.):

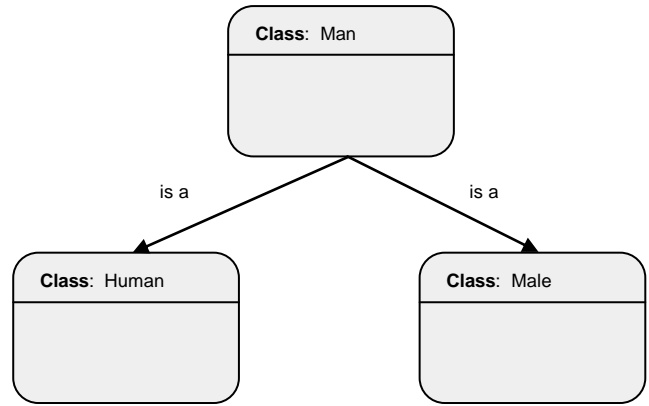

Fig. 12.Intersection of classes "Human" and "Male".

It is possible to generate the following rule:

**IF**Man**THEN** Human **and** Male, (13)

This rule means that if there is an instance of class "Man", then this instance belongs to the classes "Human" and "Male". This rule can be generated from the following OWL ontology code fragment:

```
<owl:Classrdf:ID="Man">
<owl:intersectionOfrdf:parseType="Collection">
<owl:Classrdf:about="#Male"/>
<owl:Classrdf:about="#Human"/>
</owl:intersectionOf>
</owl:Class/>
```

The following case of rule generation is when ontology has a property with a "owl:inverseof" construct. This construct is used to define an inverse relation between properties. For example, there are two classes "Human" and "Plane". There is a relation "owns" and an inverse relation "is_owned_by" between them (Fig.13.):

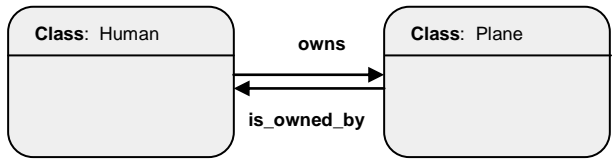

Fig. 13. "Owns" property and "is_owned_by" inverse property.

It is possible to generate such rules:

**IF** Human **THEN** (owns Plane), (14)

**IF** Plane **THEN** (is_owned_by Human), (15)

The first rule means that if there is some instance of "Human" class, then this instance has relation "owns" to class "Plane". The second rule means that if there is some instance of "Plane", then this instance has relation "is_owned_by" to class "Human". These rules can be generated from the following OWL ontology code fragment:

```
<owl:ObjectPropertyrdf:ID="owns">
<owl:inverseOfrdf:resource="#is_owned_by"/>
<rdfs:domainrdf:resource="#Human"/>
<rdfs:rangerdf:resource="#Plane"/>
</owl:ObjectProperty>
```

So, OWL ontology code fragments, which could be transformed to rules, were discussed here. It is planned that SWES will look for OWL ontologies in the Web and will develop its knowledge base [4]. That is, when needed ontology is found, SWES will look for such code fragments in the ontology and will form the knowledge base. In such a way, not simply information, but knowledge about interested domain with the possibility of inference is collected. Thus, SWES will serve as an expert system shell, which will receive a request from the users, build its own knowledge base according to the user's inquiry and render the expert help in the area of the domain [4].

## III GENERATED RULECLASSIFICATION

Presented OWL ontology code fragments as other OWL ontology code fragments, described in [3], have resulted in rules. But these rules differ from each other. In this connection it is necessary to understand the difference of generated rules and hence to classify these rules. For a start, existing rule classifications have to be investigated, because it is possible that one of existing classifications may be applied to generated rules from OWL ontologies.

In general there are several classifications of rules. One of them breaks rules up into the following categories [8]:

- Relationship;
- Recommendation;
- Directive;
- Variable;
- Uncertain;
- Meta rules.

Relationship rules are used to express a direct occurrence relationship between two events. For example, if you hear a loud sound, then the silencer is not working. Recommendation rules offer a recommendation on the basis of some known information. For instance, if it is raining, then put ona raincoat. Directive rules are like recommendations rule, but they offer a specific line of action, as opposed to the `advice' of a recommendation rule. For example, if it is raining and you do not have araincoat, then wait for the rain to stop. If the same type of rule is to be applied to multiple objects, we use variable rules, or in other words rules with variables. For instance, if X is a pupil and X's GPA>3.8 then place X on honor roll. Such rules are called pattern-matching rules. The rule is matched with known facts and different possibilities for the variables are tested, to determine the truth of the fact. Uncertain rules introduce uncertain facts into the system. The example of such rule is: if you have never won a war, then you will most probably not win this time. In this classification meta rules describe how to use other rules. For example, if you are coughing and you have chest congestion, then use the set of respiratory disease rules [8].

One more classification divides rules into three categories. These categories are the following [9]:

- Knowledge declarative rules;
- Inference procedural rules;
- Meta rules.

Knowledge declarative rules state all the facts and relationships about the problem. For instance, if inflation rate declines, then the price of silver goes down. These rules are a part of the knowledge base. Inference procedural rules advise on how to solve a problem, while certain facts are known. For example, if needed rules are not in the system, then request it from anexpert. These rules are part of the inference engine. Meta rules are necessary for making rules. Meta rules reason about which rules should be considered for firing. For example, if the rules which do not mention the current goal in their premise, and there are rules which do not mention the current goal in their premise, then the former rule should be used in preference to the latter. Meta rules specify, which rules should be considered and in which order they should be invoked [9].

There is the RuleML (Rule Markup Language) hierarchy of rules [10]. It consists of reaction rules, integrity constraints, derivationrules and facts (Fig. 14.).

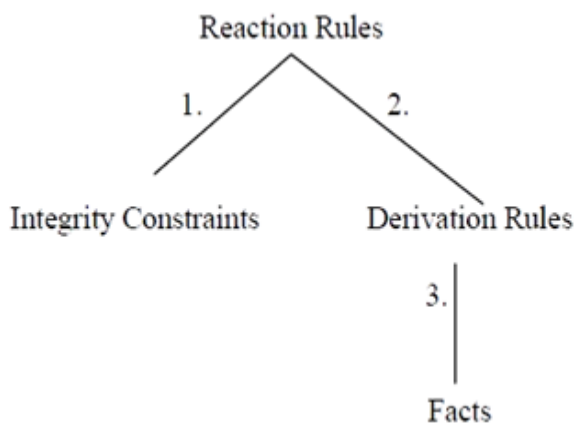

Fig. 14.The RuleML hierarchy of rules [10].

Reaction rules typically consist of an event, which starts the execution of the rule, conditions, which are necessary to execute an action, the action itself and also pre- and post- conditions. For example, a trigger in SQL (Structured Query Language) is a typical reaction rule. Integrity constraints are special reaction rules, which signal inconsistency, when certain conditions are fulfilled. An example of such rule is the following: confirmation of a booking for a room takes into account the requested room type and the requested date. Derivation rules are the rules, which assert a conclusion, when certain conditions are fulfilled. An example of such rule is: a bus is available for rental if it is not assigned to any client. Facts are special derivation rules, which have empty conjunction of conditions [10].

It can be confidently asserted that no one of listed above rule classifications is not appropriate for classification of rules, which are generated from OWL ontology. The main criterion here is the fact that using existing classifications, generated rules are distributed very irregularly. Thus, it is necessary to work out rule classification specifically for rules, obtained from ontologies. Considering all the rules, obtained from OWL ontologies, these rules may be divided into the following categories:

- Identifying rules;

- Specifying rules;

- Unobvious rules or rules, generated from hidden OWL ontology components;

- Meaning enriching rules.

The first category of rules is identifying rules. Identifying rule is the rule, which determines something, based on some characteristics. For example, rules (3), (6) are identifying rules. The second category of rules is specifying rules. Such rules are necessary to precise something, if this something is known. That is, specifying rules allow knowing the details of a particular object. For instance, rules (1), (11), (13) are specifying rules. The next category is unobvious rules or rules, generated from hidden ontology components. Hidden ontology components are components, which are not presented in ontology, but may be added based on the logic of ontology. For instance, rules (2), (4), (5), (9), (10) are unobvious rules. The last rule category is meaning enriching rules. Such rules enrich existing knowledge with new details. For example, (7), (8), (12), (14), (15) rules are meaning enriching rules.

It is necessary to note that rules are generated from OWL ontology at the same time without reference to rule category. The process of rule generation starts after merging of OWL ontologies, which are found in the Web, into a single OWL ontology. In turn, SWES looks for OWL ontologies in the Web after receiving a request from the user [5]. The process of ontology merging has two purposes. The first purpose is to obtain single and complete OWL ontology. This is made, based on technical reasons, because processing of a single OWL ontology is embedded in Jena in contrast to the processing of multiple OWL ontologies. It should be reminded that the Jena framework had been chosen for implementation as the SWES inference engine [6]. The second purpose of ontology merging is to assign the values of membership function to OWL ontology components as classes, properties, relations. The Jena framework is not designed to work with fuzzy values that is why this task and its solution will be described separately [7].

## IV CONCLUSION

This paper describes new kinds of rules, which can be generated from OWL ontology code fragments. Existing rule classifications are examined, and the original classification for generated rules is developed. Thus, this paper continues to develop the idea of OWL ontology transformation to rules. This is necessary for the Semantic Web Expert System, which will use OWL ontologies from the Web, generate rules from them and supplement its knowledge base [4].

It should be noted that the task of rule generation from OWL ontologies can be solved in different ways. The way, described in this paper and in one of the previous papers [3], is the simplest way for the solution of this task. This way is based on the OWL ontology code fragment patterns, which can be transformed to rules. Hence, this way of rule generation is a static way in the sense that only certain kinds of rules can be generated. This may not always be sufficient or acceptable. In any case the SWES has to own the way of all rule generation from OWL ontology. The idea is to utilize some other possibilities, which may be useful to solve this task, and this way will be described in one of the following papers. Future papers will be dedicated to some other tasks, which are derived from the problem of rule generation from OWL ontologies, too.

In general, the Semantic Web Expert System is close to implementation. Merging of OWL ontologies into single ontology, generation of rules from OWL ontology as well as Jena framework adaptation for fuzzy inference subroutines have already been implemented using Java programming language. There is an understanding of how to implement the task of OWL ontology search in the Web. Thus, it is necessary to work out the subsystem of communication with the user and to assemble all parts of the SWES into a single system.